\documentclass[preprint,12pt]{elsarticle}

\usepackage{xcolor}
\usepackage[bookmarksopen]{hyperref}
\usepackage{multirow}
\usepackage{subfig}

\usepackage{amssymb}
\usepackage{amsthm}
\usepackage{amsmath}

\journal{IJEPES}

\begin{document}

\begin{frontmatter}



\title{Data-Driven Stochastic AC-OPF using Gaussian Processes}


\author[inst1]{Mile Mitrovic}
\author[inst1]{Aleksandr Lukashevich}
\author[inst1]{Petr Vorobev}
\author[inst1]{Vladimir Terzija}
\author[inst2]{Semen Budenny}
\author[inst3]{Yury Maximov}
\author[inst3]{Deepjoyti Deka}

\affiliation[inst1]{organization={Center for Energy Systems and Technology},
            addressline={Skolkovo Institute of Science and Technology}, 
            city={Moscow},
            postcode={143025}, 
            country={Russia}}

\affiliation[inst2]{organization={AIRI},
            addressline={Nizhniy Susalnyy pereulok, 5}, 
            city={Moscow},
            postcode={105064}, 
            country={Russia}}

\affiliation[inst3]{organization={Theoretical Division},
            addressline={Los Alamos National Laboratory}, 
            city={Los Alamos},
            postcode={87544}, 
            state={NM},
            country={USA}}
            
\begin{abstract}
In recent years, electricity generation has been responsible for more than a quarter of the greenhouse gas emissions in the US. Integrating a significant amount of renewables into a power grid is probably the most accessible way to reduce carbon emissions from power grids and slow down climate change. Unfortunately, the most accessible renewable power sources, such as wind and solar, are highly fluctuating and thus bring a lot of uncertainty to power grid operations and challenge existing optimization and control policies. The chance-constrained alternating current (AC) optimal power flow (OPF) framework finds the minimum cost generation dispatch maintaining the power grid operations within security limits with a prescribed probability. 
Unfortunately, the AC-OPF problem's chance-constrained extension is non-convex, computationally challenging, and requires knowledge of system parameters and additional assumptions on the behavior of renewable distribution. Known linear and convex approximations to the above problems, though tractable, are too conservative for operational practice and do not consider uncertainty in system parameters. This paper presents an alternative data-driven approach based on Gaussian process (GP) regression to close this gap. The GP approach learns a simple yet non-convex data-driven approximation to the AC power flow equations that can incorporate uncertainty inputs. The latter is then used to determine the solution of CC-OPF efficiently, by accounting for both input and parameter uncertainty. The practical efficiency of the proposed approach using different approximations for GP-uncertainty propagation is illustrated over numerous IEEE test cases. 
\end{abstract}




\end{frontmatter}



\section{Introduction}
Further development of new technologies and the growing need for renewable energy sources (RES) bring significant benefits in reducing the cost of energy production, and in protecting the environment. Despite these benefits, the increase in RES generation such as wind and solar lead to significant challenges to the Transmission System Operators (TSOs)~\cite{liu2012challenges}. Instability of RES generation is probably the most important of the above challenges as it substantially compromises grid security and may lead to a power blackout. To this end, grid optimization and control algorithms that account for generation uncertainty are needed. 

This paper focuses on the Chance-Constrained Optimal Power Flow (CC-OPF)~\cite{stott2012optimal,venzke2020chance,roald2013analytical,roald2017chance,viafora2020chance} problem that minimizes energy generation cost and provides optimal power dispatch in response to changing demand and operational conditions of a grid. Further, the operational grid conditions such as voltage or transmission capacity limits are satisfied with a high probability thus guaranteeing grid immunity against a sudden change in RES generation. The chance-constrained (CC) OPF remains a hard grid optimization problem both in terms of solution accuracy and computational complexity. Monte Carlo scenario-based CC-OPF \cite{calafiore2006scenario, vrakopoulou2013probabilistic}, with nonlinear AC-PF equations, does not computationally scale for large systems or a large number of scenarios \cite{mezghani2020stochastic}. Consequently, advanced sampling policies (importance sampling, active sampling) have been recommended to reduce the number of samples necessary \cite{capitanescu2012cautious,owen2019importance,mezghani2020stochastic,lukashevich2021importance,lukashevich2021power}. In contrast, analytical approach to CC-OPF states the chance constraints using distributional information of the uncertainty and is the focus of this work. A substantial number of papers is devoted to designing convex/linear approximations to the chance-constraints problem~\cite{bienstock2014chance,lubin2019chance,lubin2015robust,roald2016corrective,du2021chance}; however, most of these formulations are conservative and hence are unable to address a large change in power generation. Another line of research is focused on the nonlinear Polynomial chaos expansion for modeling uncertainties in power demand and generation~\cite{muhlpfordt2019chance,muhlpfordt2016solving}. Although this approach often gives acceptable accuracy, its solubility lacks good computational performance. Furthermore, most prior work requires knowledge of the power flow model parameters that may not be accurate due to insufficient and irregular equipment calibration. Moving forward to the realm of micro-grids and low-voltage grids, such parametric uncertainties will be major due to reduced real-time metering. 

This paper proposes a flexible alternative to the traditional approaches for solving the CC-OPF problem that balances the computational complexity and accuracy of the solution. In particular, having empirical samples of RES and load fluctuations, we propose to learn a Gaussian process-based approximation to the AC power flow balance equations and use them inside CC-OPF. Gaussian process (GP) is a well-known model-free machine learning algorithm that can fit well any smooth nonlinear function~\cite{dudley2010sample} and yields a suitable complexity-accuracy tradeoff~\cite{schulz2018tutorial,liu2020gaussian}. GP has a competitive advantage against traditional approximation algorithms that are focused on fitting the power balance equations around the current state and, therefore, fail to capture non-linearity from large fluctuations. In prior work, GP and neural network-based power-flow models have been implemented in deterministic OPF formulations~\cite{pareek2020gaussian, zamzam2020learning, pan2020deepopf,kekatos}; however, such models do not consider CC-OPF with input uncertainty. Neural network-based stochastic AC-OPF has been proposed in \cite{gupta2021dnn}, but considers scenario-based chance constraints. In contrast, our GP-based formulation exploits uncertainty propagation in Gaussian Processes \cite{hewing2019cautious}. It includes analytical CC-OPF constraints that account for both parametric, i.e., line impedances, and input, i.e., RES, loads, uncertainties. 

Our case studies over various IEEE test cases demonstrate the better or comparable performance of Gaussian processes based stochastic AC-OPF over the scenario and model-based CC-OPF approaches, despite not having access to network parameters.

The paper is organized as follows. Section \ref{sec:00-background} introduces the problem, summarizes related works and paper notation. Section \ref{sec:10-GP} outlines Gaussian process background. The algorithm for solving the chance constrained AC-OPF problem is introduced in Section~\ref{sec:20-Algo}. The results of the case study are given in Section~\ref{sec:30-Case}. Sections \ref{sec:40-Discussion} and \ref{sec:50-Conclusion} contain a critical discussion of the approach, its possible extensions and applications respectively.

\section{Problem Setup and Notation}\label{sec:00-background}
This section covers the notation used throughout the paper and introduces the setup of the chance-constrained AC optimal power flow problem considered in this paper.

\begin{table}[ht]
 \centering
 \resizebox{\textwidth}{!}{\begin{tabular}{l|l|l|l}
 $\alpha$ & participation factor & $g(\cdot) = 0$ & AC power flow equations \\
 $w$ & random independent variable (fluctuation) & 
 $\Omega$ & total active power imbalance \\
 $n_u$ & number of controllable generators  & $n_L$ & number of loads \\
 $n_R$ & number of renewables & $\gamma$ & power ratio \\
 $p(\cdot), E[\cdot]$ & probability, expectation & $\texttt{Prob}(\cdot)$ & chance (probability) constraint \\
 $\varepsilon$ & acceptable violation probability & $\varrho$ & grid loss percentage \\
 $\mathcal{B}$ & set of buses & $\psi$ & uniform distribution \\
 $\mathcal{E}$ & set of lines & $x$ & vector of input features \\
 $G_{bk}/B_{bk}$ & real$/$imaginary part of admittance of line $bk$ & $\mathcal{N}(\mu, \Sigma)$ & Gaussian distribution \\
 $p/q$ & active$/$reactive power injections & $diag(\cdot)$ & diagonal matrix \\
 $v/\theta$ & voltage magnitudes$/$phase angels & $tr(\cdot)$  & trace of a square matrix \\
 $s$ & apparent power flow & $k$ & squared exponential kernel \\
 $\mathcal{B}_c$ & set of controllable generators & $K$ & Gram matrix  \\
 $c2, c1, c0$ & cost coefficients & $I$ & identity matrix \\
 $d$ & vector of random independent variables & $m(\cdot)$ & prior mean \\
 $u$ & vector of controllable variables & $r$ & quantile function of ${\cal N}(0,1)$ \\
 $y$ & vector of output dependent variables & $N$ & number of training samples \\
 $\eta/\nu$ & log-normal distributions of renewables &&\\
 \end{tabular}}
 \caption{Paper notation}
 \label{tab:notation}
\end{table}

\subsection{Notation}
For the given matrix $X$, $X_{ij}$ denotes element $ij$ of a matrix $X$, and $[x]_i$, $[x]_{:,j}$ its $i$-th row and $j$-th column, respectively. With $[x]_{*,j}$, we denote $*$ element of $j$-th column vector. Further, we will use $\cdot_*$ subscript to denote unseen test data. For the given vector $x$, $diag(x)$ is used to refer to a diagonal matrix with entries of vector $x$. The probability density of $x$ is denotes as $p(x)$, $p(x|y)$ representing the probability of $x$ for given $y$. $\frac{\partial \cdot (x_*)}{\partial x_*}$ is partial derivative of function $\cdot$ evaluated at~$x_*$. The power grid related notation is summarized in Table \ref{tab:notation}.

\subsection{AC-OPF}
Let $\mathcal{G}=(\mathcal{B},\mathcal{E})$ be a transmission power grid given by its buses $b \in \mathcal{B}$ and lines $e_{bk} \in \mathcal{E}$\footnote{We also use $b\sim k$ to represent that $b,k$ share an edge.}. The alternating current power flow (PF) equations are then
\begin{subequations}\label{eq:pf}
\begin{align}
 p_{bk} = v_bv_k \left(G_{bk}\cos\theta_{bk}+B_{bk}\sin\theta_{bk}\right) \\
 q_{bk} = v_bv_k\left(G_{bk}\sin\theta_{bk}-B_{bk}\cos\theta_{bk}\right)
\end{align}
\end{subequations}
where $p_{bk}$ and $q_{bk}$ are the active and reactive power flows from bus $b$ to bus $k$ along $e_{bk}$ line; $\theta_{bk}:=\theta_b-\theta_k$ is a phase angle difference between adjacent buses $k$ and $b$, $k\sim b$; $G_{bk}$ and $B_{bk}$ are the the real and imaginary parts of the admittance on the line between $b$ and $k$. The system is balanced when the power flows
leaving each bus is equal to the network injection at that bus:
\begin{subequations}\label{eq:pf_inj}
\begin{align}
 p_b = v^2_b G_{bb} + \sum_{k\sim b} p_{bk} \\
 q_b = -v^2_b B_{bb} + \sum_{k\sim b} q_{bk},
\end{align}
\end{subequations}
where the first term represents nodal shunt elements; $p_b$ and $q_b$ are active and reactive power injections at bus $b$. To simplify the notation we refer the PF Eqs.~\eqref{eq:pf},\eqref{eq:pf_inj} as $g(p, q, v, \theta) = 0$.

The AC-OPF problem minimizes the cost of generation while satisfying the load and system constraints, as formulated below. 
\begin{subequations}\label{eq:opf}
\begin{align}
 &\min_{p_g} 
 f(p_g) 
 \\
 &\text{s.t.~} g(p, q, v, \theta) = 0, \\
 &~~~~~ p_b^{\min} \leq p_b \leq p_b^{\max}, \quad \forall b\in {\cal B}_c\label{eq:00}\\
 &~~~~~ q_{b}^{\min} \leq q_{b} \leq q_{b}^{\max}, \quad \forall b\in {\cal B}_c\\
 &~~~~~ v_{b}^{\min} \leq v_{b}\leq v_{b}^{\max}, \quad \forall j\in \mathcal{B}\\
 &~~~~~ s_{bk}(w) \leq s_{bk}^{\max}, ~~~~~~~ \forall b\sim k\label{eq:10}
\end{align}
\end{subequations}
where $p_g$ is the controllable power generations. The voltage at generators is fixed at their prescribed values. ${\cal B}_c$ is a set of controllable buses, and $s_{bk}:=\sqrt{p_{bk}^2+q_{bk}^2}$ is a value of apparent power on the line $e_{bk}$ that is limited from above by $s_{bk}^{\max}$ to prevent extreme currents and line overheating. Problem~\eqref{eq:opf} can be equally stated as:
\begin{subequations}\label{eq:opf2}
\begin{align}
 &~~~~~\min_{u} f(u) \\
 &~\text{s.t.~~} g(y, u, d) = 0 \\
 &~~~~~~ h^{min} \leq h(y, u, d) \leq h^{max} \label{4c}
\end{align}
\end{subequations}
where control variable $u = p_g$ is bounded in a range $u \in [u^{\min}; u^{\max}]$; $d$ is a vector of input variables (active power on loads $p_l, q_l$ and RES $p_{rs}, q_{rs}$). The reactive power at loads $q_l$ and RES $q_{rs}$ are implicitly included under a constant power factor $cos(\phi)$ assumption, i.e., $q_{l} = \gamma p_{l}$, $q_{rs} = \gamma p_{rs}$, where power ratio $\gamma = \frac{\sin(\phi)}{\cos(\phi)}$. 

Let $y$ be a vector of (dependent) output variables given by the solution of \eqref{eq:pf},\eqref{eq:pf_inj}; and $h$ is a set of constraints given by Ineq.~\eqref{eq:00}--\eqref{eq:10}. Usually, the dependent constrained variables include the voltage magnitude $v$ at PQ (load) and slack busses, reactive power $q$ at PV (generator) buses, and apparent power flows $s$ on the lines.

\subsection{Chance Constrained AC Optimal Power Flow}\label{sec:cc_opfmodel}
Deviations in loads and renewables are caused due to forecast errors, external fluctuations, or intra-day electricity trading. Therefore, we propose that both load $d_{l} = [\bar{p}_{l}, \bar{q}_{l}]$ and RES $d_{rs} = [\bar{p}_{rs}, \bar{q}_{rs}]$ as uncertain power injections. We assume that active power uncertainties are modeled as the sum of the forecast set points $p^0_d$ and a fluctuation $\omega \sim {\cal W}$\footnote{$W$ is the uncertainty distribution.} as,
\begin{center}
 $p_d(\omega) = p^0_{d} + \omega$,\\
\end{center}
where $\omega$ is an independent random vector with zero mean and known standard deviation $\sigma_{\omega}$. The reactive power fluctuations are modeled with constant power factor as mentioned before as, $q_d(\omega) = \gamma p_d(\omega) = \gamma p^0_d + \gamma \omega$. Thus, the ratio of active and reactive power injections at each node remains unchanged during fluctuations. 
 
Designing power dispatch $p_g(\omega)$ that is absolutely immune against any uncertainty realization leads to an infeasible or a very conservative solution with limited impact in practice. The chance-constrained formulation, which admits a small probability for each of the inequality constraints not being feasible, allows deriving practically methods having rigorous theoretical guarantees~\cite{liu2009theory}. The chance-constrained AC optimal power flow problem is given as follows for $\omega \sim W$:
\begin{subequations}\label{eq:opf3}
\begin{align}
 &~~~~~\min_{u} \mathbb{E}_\omega f(u(\omega)) \\
 &~\text{s.t.~~} g(y,u,d) = 0 \\
 &~~~~~~ \begin{cases}
 \texttt{Prob}_{\omega \sim W}(h(y,u,d)) \leq h^{\max}) \geq 1-\varepsilon_y\\
 \texttt{Prob}_{\omega \sim W}(h(y,u,d)) \geq h^{\min}) \geq 1-\varepsilon_y
 \label{eq:cc00}\\
 \texttt{Prob}_{\omega \sim W}(u(\omega) \leq u^{\max}) \geq 1-\varepsilon_u\\
 \texttt{Prob}_{\omega \sim W}(u(\omega) \geq u^{\min}) \geq 1-\varepsilon_u
 \end{cases}
\end{align}
\end{subequations}
where the dependence on $\omega$ for $g$ and $h$ is suppressed. Also, $\varepsilon_y$ and $\varepsilon_u$ correspond to probabilities of the constraint violation that are taken according to the measure $W$ of uncertainty in power demands and RES generation. 

\subsection{Automated Generation Control}
In the rest of the paper, we maintain the power balance and AC OPF security constraints by changing the active power generation only. The Automatic Generation Control (AGC)~\cite{apostolopoulou2014automatic}, which is widely accepted in modern power systems, stands for compensating the power injections mismatch $\Omega = \sum_{b=1}^{n} \omega_b$ by proportionally changing selected power generations in set ${\cal B}_c$ according to the vector of non-negative participation factors $\{\alpha_b\}$
\begin{gather}\label{feedback}
 u_b(\omega) = u_b + \alpha_b \Omega, \quad b \in {\cal B}_{c}, \quad \sum_{b\in {\cal B}_{c}} \alpha_b = 1,
\end{gather}
The parameters $\{\alpha_b\}$ can be optimized along with the generator set points $p_g^0$ to define the corrective control efficiency. 

\subsection{Data Driven  AC Power Flow Optimization Pipeline}

In this paper, we use the following pipeline for solving the chance constrained AC optimal power flow problem using a data-driven approach. 

\begin{figure}[ht]
\centering
\includegraphics[width=0.90\textwidth]{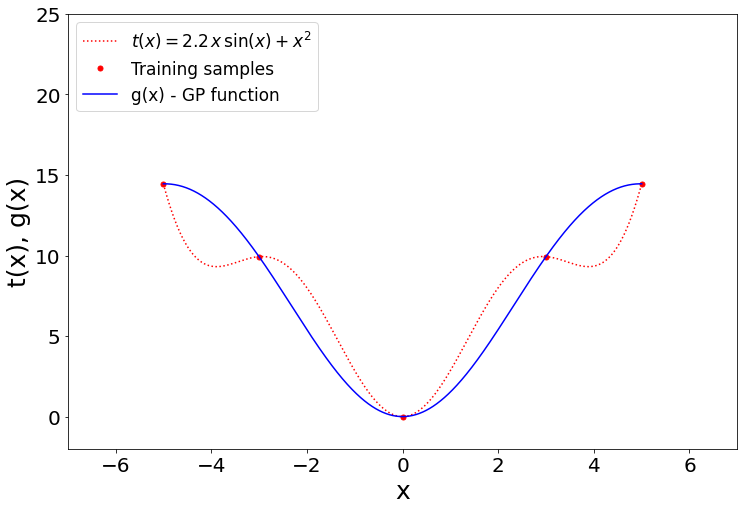}
\caption{A synthetic example illustrates the power of the Gaussian process regression. Function $t(x) = 2.2 x\sin x + x^2$ stands for the initial equation and $g(x)$ is its data-driven approximation via Gaussian process with squared exponential kernel. We refer to the approximation as a GP function.}
\label{fig:00}
\end{figure}

In the first stage, we collect fluctuations of power demand and RES power generation and dependent variables using historical data or a simulator. We use this data and the Gaussian process regression in the second stage to provide a data-driven probabilistic approximation for the dependent variables in the power balance equations, Eqs.~\eqref{eq:pf},\eqref{eq:pf_inj}. The approximation is substantially more tractable than the PF equations and leads to a simpler optimization problem. Figure~\ref{fig:00} illustrates the above for a synthetic example. 

In the third stage, we compute the probability for constraint violation (termed \emph{uncertainty margins}) that are given by Eqs.~\eqref{eq:cc00}. This probability has an explicit expression for Gaussian process-based modeling with input uncertainty and enables the CC-OPF problem to be cast as a deterministic optimization. Solving the latter gives us the minimal cost of power dispatch, satisfying the probabilistic security constraints. 

In the next section, we introduce the data-driven Gaussian process regression framework used throughout the paper in our CC-OPF formulation.

\section{Gaussian Processes}\label{sec:10-GP}

\subsection{Gaussian Process Regression}
Gaussian process regression (GPR) is a data-driven non-parametric regression approach that balances approximation accuracy and complexity. An attractive property of GPR is that it provides second-order statistics for the predicted variables. We employ Gaussian process regression to infer the power flow function $g(y,u,d)$ from collected input and output measurement samples [$(x_i, y_i), i=1,...,N$]. To reiterate, each input data sample $x_i\in \mathbb{R}^{n_x}$ includes the vector of controllable active power generators $u = p_g$, and active power injections $p^\top_d = [p^\top_l~p^\top_{rs}]$. Thus, $x^\top_i = [u_i^\top, p_{d_i}^\top]$. The output samples $y_i\in \mathbb{R}^{n_y}$ corresponds to all dependent variables involved in constraint \eqref{eq:opf2}, and $y^\top_i=[v_i^\top, q_i^\top, s_i^\top]$. For the quasi-stationary power system, the output samples can be considered as outputs from the nonlinear AC-PF equation as:
\begin{gather}\label{eq:output}
 y_i = g\left(x_i, q_{l,i}, q_{rs,i}\right) + \zeta_i, \quad i=1,\dots,N 
\end{gather}
where $N$ is total number of samples and $\zeta_i \sim \mathcal{N}(0, \Sigma_{\zeta})$ is the measurement Gaussian noise/modeling error with covariance $\Sigma_{\zeta}$. 

Here we want to note that the reactive power at $q_l$ (loads) and $g_{rs}$ (RES) are not considered as inputs to the GP model as they are given by a fixed power factor ratio from the active power. By excluding redundant features, we reduce computational complexity without compromising the model performance.

For the given input data matrix $X = [x_i^\top]_{1\leq i\leq N} \in \mathbb{R}^{N \times n_x}$, each output dimension $[y]_{:,a} \in \mathbb{R}^N$ is learnt independently. A GP \cite{rasmussen2003gaussian} with prior mean function $m_a(\cdot)$ and chosen kernel $k_{a}(\cdot,\cdot)$ in each output dimension $a \in \{1,...,n_y\}$, models the output $[y]_{:,a}$ as normally distributed data:
\begin{gather*}
 [y]_{:,a} \sim \mathcal{N}(m_a(X), K_a + \sigma^2_{\zeta,a}I )
\end{gather*}
where $\sigma^2_{\zeta, a}$ is the covariance of measurement noise $\zeta$ in \eqref{eq:output}; $I$ is identity matrix; $m_a(X) = [m_a(x_1),...,m_a(x_N)]^\top$, and $[K_a]_{ij} = k_a(x_i,x_j)$ is the Gram matrix of the sampled data points. In this paper, we consider a zero prior mean $m(X)=0$ and the squared exponential (SE) kernel with Automatic Relevance Determination (ARD):
\begin{gather*}
 k_a(x_i,x_j) = \sigma^2_{f,a}\exp\left(-\frac{1}{2}(x_i-x_j)^\top \Lambda^{-1}_a (x_i-x_j)\right)
\end{gather*}
in which $\sigma^2_{f,a}$ is the signal variance and $\Lambda_a = diag[l^2_{a,1}, l^2_{a,2},..., l^2_{a,{n_x}}]$ is a positive diagonal length scale matrix. 

\subsubsection{Hyperparameters learning}
The parameters of GP to be optimized are the aforementioned hyperparameters $\Theta = [\Lambda_a, \sigma^2_{f,a}, \sigma^2_{\zeta,a}]$ of the covariance function $K_a$ and process noise. Note that $\Lambda_a$ is a diagonal matrix. We employ the Maximum Likelihood Estimate (MLE) with the log marginal likelihood to optimize hyperparameters as follows:
\begin{align}\label{eq:mle} 
 \log~p([y]_{:,a}|X, \Theta) = \frac{1}{2}[y]_{:,a}^{T}(K_a & + \sigma^2_{\zeta,a}I)[y]_{:,a} \nonumber\\
 & - \frac{1}{2}\log|K_a + \sigma^2_{\zeta,a}I| - \frac{N}{2}\log 2 \pi
\end{align}
The negative log marginal likelihood (NLL) is applied as the objective function to minimize the function value subject to the hyperparameters. The nonlinear optimization solver was employed since the NLL is a non-convex function. This paper uses Sequential Least Squares Programming (SLSQP) solver to find the optimal solution.

\subsubsection{Prediction with Deterministic Input}
For a given unseen sample data $x_*$, the predictive distribution of $[y]_{*,a}$ is provided as a marginal distribution with zero mean:
\begin{gather*}
 [y]_{*,a} \sim \mathcal{N}(0, K_{**})
\end{gather*}
Thus, a joint distribution of the training data and unseen point $x_*$ is given by:
\begin{gather*}
 p([y]_{*,a}, [y]_{:,a}) \sim \mathcal{N}\left(\begin{bmatrix}
 0\\
 0
 \end{bmatrix}, 
 \begin{bmatrix}
 K_a + \sigma^2_{\zeta,a}I & k_*\\
 k^\top_* & K_{**}
 \end{bmatrix}\right)
\end{gather*}
where $k_* = k_a(x_i, x_*)$ is the vector of covariance between unseen point and training samples and $K_{**} = k_a(x_*, x_*) = 1$ is scalar. Since the joint distribution of the variables being Gaussian, the resulting distribution conditioned on the training data $p([y]_{*,a}|[y]_{:,a}) = \mathcal{N}(\mu_a(x_*), \sigma^2_a(x_*))$ is also Gaussian \cite{rasmussen2003gaussian} with:
\begin{subequations}\label{eq:gp_00}
\begin{align}
 \mu_a(x_*) &= k^\top_*(K_a + \sigma^2_{\zeta,a}I)^{-1}[y]_{:,a} \label{eq:gp_01} \\
 \sigma^2_a(x_*) & = K_{**} - k^\top_*(K_a + \sigma^2_{\zeta,a}I)^{-1}k_* \label{eq:gp_02}
 \end{align}
\end{subequations}

Generally, the predictive distribution of multivariate output dimension $y_*$ can be given by the equations:
\begin{align*}
 \mu(x_*) & = [\mu_1(x_*), \mu_2(x_*), ..., \mu_{n_y}(x_*)]^\top \\
 \Sigma(x_*) & = diag([\sigma^2_1(x_*), \sigma^2_2(x_*), ..., \sigma^2_{n_y}(x_*)])
\end{align*}

To minimize the computational effort in inverting $(K_a + \sigma^2_{n,a}I)$ during prediction, we pre-compute it by applying the Cholesky decomposition. 

\subsection{Uncertainty Propagation}\label{sec:uncertainty}
So far, we have assumed that the input to the GP is deterministic and the output is Gaussian with errors due to measurement/modeling errors. For CC-OPF, we need to consider uncertainty in input forecasts propagating through the learned GP model. Following standard convention in prior work, we model all uncertain inputs have a Gaussian distribution $x_* = \mathcal{N}(\mu_{x_*}, \Sigma_{x_*})$. The distribution of the predicted uncertain output for a particular dimension $a$ is obtained by integrating over the input distribution as follows:
 \begin{gather*}
 p(y_* | \mu_{x_*}, \Sigma_{x_*}) = \int p(y_*|x_*)\, p(x_*| \mu_{x_*}, \Sigma_{x_*}) \,dx_* 
 \end{gather*}
This posterior distribution cannot be computed analytically for general Kernel functions since the Gaussian input is mapped through a nonlinear function. To overcome this issue, we approximate the predictive distribution by a Gaussian and use different methods to derive its statistics, thus propagating the uncertainty. In the following, we gave a brief overview of Taylor Approximation (TA) and Exact Moment Matching (EM) that we consider in our setting. For details we refer the reader to~\cite{girard2003gaussian, deisenroth2010efficient}.

\subsubsection{Taylor Approximation}
We approximate output mean $\mu(x_{*})$ by its first-order Taylor expansion and output variance $\sigma^2(x_{*})$ by its 1st and 2nd order Taylor expansion around $\mu_{x_*}$. With the first-order Taylor expansion, the mean prediction at a
random input $x_*$ does not provide any correction over the zero-order, as in \eqref{eq:gp_01}:
\begin{gather}\label{eq:mean_TA1}
 \mu_{TA}(x_*) = k^\top_*(\mu_{x_*})(K + \sigma^2_n I)^{-1}y
\end{gather}

For output variance, we consider the case with the first order Taylor approximation (TA1) as follows:
\begin{gather*}
 \sigma^2_{TA1}(x_*) = \sigma^2(\mu_{x_*}) + \frac{\partial \mu_{TA}(\mu_{x_*})}{\partial \mu_{x_*}}^\top \Sigma_{x_*} \frac{\partial \mu_{TA}(\mu_{x_*})}{\partial \mu_{x_*}} 
\end{gather*}
as well as the second order Taylor approximation (TA2):
\begin{gather*}
 \sigma^2_{TA2}(x_*) = \sigma^2_{TA1}(x_*) + \frac{1}{2}tr\left(\frac{\partial^2 \sigma^2(\mu_{x_*})}{\partial \mu_{x_*} \partial\mu_{x_*}^\top}\Sigma_{x_*}\right)
\end{gather*}
where $\sigma^2(\mu_{x_*})$ corresponds to \eqref{eq:gp_02}.

\subsubsection{Exact Moment Matching}
According to~\cite{deisenroth2010efficient}, for a zero prior mean function and SE covariance kernel with ARD, the predicted mean and variance could be analytically computed by the law of iterated expectations (Fubini's theorem). Thus, the predicted mean is given by:
\begin{align}
 \mu_{EM}(x_*) = k^\top_*(\mu_{x_*}, \Sigma_{x_*})(K + \sigma^2_n I)^{-1}y
\end{align}
where the kernel $k_*^\top$ is simultaneously a function of input mean and covariance:
\begin{gather*}
  k^\top_*(\mu_{x_*}, \Sigma_{x_*}) = \sigma_f^{2}|\Sigma_{*}\Lambda^{-1} + I|^{-\frac{1}{2}} \exp\left\{-\frac{1}{2}(x-\mu_{*})^\top(\Sigma_{*} + \Lambda)^{-1}(x-\mu_{*})\right\} 
\end{gather*}

The variance of the predictive distribution is obtained as follows:
\begin{gather*}
 \sigma^2_{EM}(x_*) = \sigma_f^2 - tr\left((K + \sigma^2_n I)^{-1}Q\right) + \beta^\top Q\beta - \mu_{x_*}^2
\end{gather*}
in which $\beta = (K + \sigma^2_n I)^{-1}y$ and elements of $Q \in \mathbb{R}^{N\mathsf{x}N}$ are computed as:
\begin{align*}
 [Q]_{ij} = & \frac{k_*(x_i,\mu_{x_*}) k_*(x_j,\mu_{x_*})}{|2\Sigma_{x_*}\Lambda^{-1} + I|^{\frac{1}{2}}} \cdot \\
 & \qquad\qquad  \exp\left((z_{ij}-\mu_{x_*})^\top\left(\Sigma_{x_*}+\frac{1}{2}\Lambda\right)^{-1}\Sigma_{x_*}\Lambda^{-1}(z_{ij} - \mu_{x_*})\right)
\end{align*}
with $z_{ij} = \frac{1}{2}(x_i + x_j)$.
\section{Algorithm}\label{sec:20-Algo}

\subsection{GP CC-OPF Formulation}
The model given by Eq.~\eqref{eq:opf3} is reformulated as a nonlinear programming (NLP) problem with learned GPs in the constraints, such that an optimization solver can handle it.

Following~\cite{bienstock2014chance}, we reformulate the objective function as:
\begin{gather}\label{eq:obj} 
 E_{\omega}[f(\cdot)] = \sum_{k\in n_u} \left\{ c_{2k}(u_{k}^{2} + tr(\Sigma_w)\alpha_{k}^{2}) + c_{1k}u_k + c_{0k}\right\}
\end{gather}
where $\{c_{2k}, c_{1k}, c_{0k}\}_{k=1}^{n_u} \geq 0 $ are scalar cost coefficients and $\alpha$ is the participation factor. Note that $\alpha$ (participation factor) is not a standard (fixed) value, but a variables to be optimized with constraints $\sum_{k=1}^{n_u}\alpha_k=1$ and $\alpha_k \geq 0$ . Accordingly, the objective function is a convex quadratic function of $u$ and $\alpha$.

Reformulating chance constraints to the tractable form implies reformulating both output and decision input constraints. According to~\cite{hewing2019cautious}, a GP individual output probabilities, could be expressed in terms of the mean $\mu_y$
and variance $\sigma^2_y$:
\begin{gather}\label{eq:pr_y}
 \begin{cases}
 \texttt{Prob}(\mu_{y_z} + \sqrt{\sigma^2_{y_z}} \leq y_z^{\max}) \leq 1-\varepsilon_{y_z}\\
 \texttt{Prob}(\mu_{y_z} - \sqrt{\sigma^2_{y_z}} \geq y_z^{\min}) \leq 1-\varepsilon_{y_z}\\
 \end{cases}
\end{gather}
that is equivalent to:
\begin{gather}\label{eq:pr_y_1}
 y_z^{\min} + r_{y_z} \sqrt{\sigma^2_{y_z}} \leq \mu_{y_z} \leq y_z^{\max} - r_{y_z} \sqrt{\sigma^2_{y_z}}
\end{gather}
where $r_{y_z}$ represents the quantile function $\Phi^{-1}(1-\varepsilon_{y_z})$ of the standard normal distribution and $z=1,...,n_y$.

Similarly, the decision input constraints are~\cite{bienstock2014chance}:
\begin{gather*}
 \begin{cases}
 \texttt{Prob}(u_k + \alpha_k\, tr(\Sigma_w)\leq u_k^{max}) \leq 1-\varepsilon_{u_k}\\
 \texttt{Prob}(u_k - \alpha_k\, tr(\Sigma_w) \geq u_k^{min}) \leq 1-\varepsilon_{u_k}
 \end{cases}
\end{gather*}
As in \eqref{eq:pr_y}, this formulation is equivalent to:
\begin{gather}\label{eq:pr_u_1}
 u_k^{min} + r_{u_k} \alpha_k\, tr(\Sigma_w) \leq u_k \leq u_k^{max} - r_{u_k} \alpha_k\, tr(\Sigma_w)
\end{gather}
where $r_{u_k} = \Phi^{-1}(1-\varepsilon_{u_k})$ and $k=1,...,n_u$.

By taking into account of the formulations in \eqref{eq:obj}, \eqref{eq:pr_y_1}, and \eqref{eq:pr_u_1}, the following approximate GP CC-OPF is derived:
\begin{subequations}\label{eq:opf4}
\begin{align}
 &\min_{u, \alpha, \mu_y, \sigma_y} \sum_{k\in n_u} \{ c_{2k}(u_{k}^{2} + tr(\Sigma_w)\alpha_{k}^{2}) + c_{1k}u_k + c_{0k}\} \label{eq:Gp-cost}\\
 &\text{s.t.~} \sum_{k \in n_u} \alpha_k = 1, ~ \alpha_k \geq 0 \\
 &~~~~~ \sum_{k\in n_u} u_k = \sum_{i\in n_L} p_{l_i} - \sum_{j\in n_R} p_{rs_j} \label{eq:balance} \\
 &~~~~~~ \mu_y = \mu(x_*) \label{eq:balance_mean}\\ 
 &~~~~~~ \sigma_y^{2} = \sigma^{2}(x_*) \label{eq:balance_var}\\
 &~~~~~~ y_z^{min} + \lambda_{y_z} \leq \mu_{y_z} \leq y_z^{max} - \lambda_{y_z}\label{eq:const_var1}\\
 &~~~~~~ u_k^{min} + \lambda_{u_k} \leq u_k \leq u_k^{max} - \lambda_{u_k} \label{eq:const_var2}
\end{align}
\end{subequations}
where $\lambda_{y_z} = r_{y_z} \sqrt{\sigma^2_{y_z}}$ for $z\in n_y$ and $\lambda_{u_k} = r_{u_k} \alpha_k\, \sqrt{tr(\Sigma_w)}$ are uncertainty margins. By $\sqrt{tr(\Sigma_w)}$ the standard deviation of the total active power imbalance $\Omega$ is represented. Equation \eqref{eq:balance} describes the balance equation in which decision variables of controllable generators have to be satisfied in the power system keeping the balance between generation and consumption, while \eqref{eq:balance_mean} and \eqref{eq:balance_var} represent GP output mean and variance formulation. Thus, we replace the standard AC power flow equation with a data-driven method using these equations.

The proposed OPF optimization method with GP-based model results in a non-convex optimization problem which is often hard to solve. The second-order derivative information of all quantities is available for the chosen twice differentiable SE kernel and zero prior means. Thus, the problem can be solved using sequential quadratic programming or the nonlinear interior-point method. We employed a primal-dual interior-point algorithm with filter line-search. The available IPOPT~\cite{wachter2006implementation} solver is used in non-linear optimization framework CasADi~\cite{andersson2019casadi}. To solve this optimization problem, we assume simultaneous approaches that simultaneously treat $u$, $\alpha$, $\mu_y$, and $\sigma^2_y$ as optimization variables.

\subsection{Input Covariance}
As previously mentioned in Section \ref{sec:uncertainty}, we assume that unseen input points $x_*$ for the GP to be normally distributed with specified vector mean $\mu_{x_*} \in \mathbb{R}^{n_x}$ and covariance matrix $\Sigma_{x_*} \in \mathbb{R}^{n_x\mathsf{x} n_x}$. Mean values correspond to optimized control variables and forecast uncertain parameters as:
\begin{gather*} 
 \mu_{x_*} = [u^\top, p_{l}^\top, p_{rs}^\top]^\top
\end{gather*}
where $u=[p^1_g, p^2_g, ..., p^{n_u}_g]^\top$, $p_l=[p^1_l, p^2_l, ..., p^{n_L}_l]^\top$ and $p_{rs}=[p^1_{rs}, p^2_{rs}, ..., p^{n_R}_{rs}]^\top$.
We need to correctly define the input covariance matrix $\Sigma_{x_*}$ as $u$ is correlated with uncertainty $w$ in $p_l$ and $p_{rs}$ due to the linear feedback, see Eq.~\eqref{feedback}. This matrix is:
\begin{equation} 
 \Sigma_{x_*} =
 \begin{bmatrix}
 \Sigma_u & \Sigma_{uw}\\\
 \Sigma_{uw}^\top & \Sigma_w
 \end{bmatrix}
\end{equation}
so that $\Sigma_u \in \mathbb{R}^{n_u \mathsf{x} n_u}$, $\Sigma_w \in \mathbb{R}^{n_d \mathsf{x} n_d}$ and $\Sigma_{uw} \in \mathbb{R}^{n_u \mathsf{x} n_d}$ are covariance sub-matrices.

The covariance sub-matrix $\Sigma_w$ represents uncertain variations in loads and renewable sources. We consider this matrix as diagonal:
\begin{gather*}
 \Sigma_{w} =
 \begin{bmatrix}
 \sigma^2_{l_1} & ... & 0 & 0 & ... & 0 \\
 ... & ... & ... & ... & ... & ...\\
 0 & ... & \sigma^2_{l_L} & 0 & ... & 0\\
 0 & ... & 0 & \sigma^2_{rs_1} & ... & 0 \\
 ... & ... & ... & ... & ... & ...\\
 0 & ... & 0 & 0 & ... & \sigma^2_{rs_R} \\
 \end{bmatrix}
\end{gather*}
and assume that fluctuations between stochastic power injections are independent, consistent with prior work \cite{bienstock2014chance}. 

Sub-matrix $\Sigma_u$ refers to controllable generator deviations caused by loads and renewable sources fluctuations keeping the balance of the system. Since controllable variables $u$ are dependent on forecasted uncertain injections $p_l$ and $p_{rs}$, the deviations of $u$ are also dependent on fluctuations of $w$ as follows:


$$ \Sigma_{u} =
 \begin{bmatrix}
 \alpha^2_1 tr(\Sigma_w) & ... & 
 \alpha_1 tr(\Sigma_w) \alpha_{n_u}\\
 ... & ... & ... \\
 \alpha_{n_u} tr(\Sigma_w) \alpha_1 & ... & \alpha^2_{n_u} tr(\Sigma_w)\\
 \end{bmatrix}$$

The covariance elements between dependent decision variables and stochastic parameters are expressed in sub-matrix $\Sigma_{uw}$ as:
$$ \Sigma_{uw} =
 \begin{bmatrix}
 \alpha_1\sigma^2_{l_1} & ... & \alpha_1\sigma^2_{l_{n_L}} & \alpha_1\sigma^2_{rs_1} & ... & \alpha_1\sigma^2_{rs_{n_R}} \\
 ... & ... & ... & ... & ... & ... \\
 \alpha_{n_u}\sigma^2_{l_1} & ... & \alpha_{n_u}\sigma^2_{l_{n_L}} & \alpha_{n_u}\sigma^2_{rs_1} & ... & \alpha_{n_u}\sigma^2_{rs_{n_R}} \\
 \end{bmatrix}$$
\section{Training Dataset Simulation}\label{sec:25-Simulation}
As our GP-based data-driven CC-OPF is model-free and uses estimated relation between input and output variables, we next discuss the creation of a synthetic dataset to learn the model and for our validation. In practice, historical data or a power grid simulator can be used to generate such data for learning. 

Motivated by~\cite{donnot2019deep}, we generated the synthetic dataset using three different parts of injection sampling, first active and reactive power loads, then renewable sources, and finally active power (controllable) generation.

\subsubsection{Sampling active power loads and renewable sources}

To simulate real active powers for $i$-th load and $j$-th renewable source, we define stochasticity as:
\begin{gather*}
 p_l^i = \eta^i p_{l_{ref}}^i, ~\forall 1\leq i \leq n_L, \;
 p_{rs}^j = \nu^j p_{rs_{ref}}^j, \; \forall 1\leq j \leq n_R 
\end{gather*}
where $p_{l_{ref}}^i$ and $p_{rs_{ref}}^j$ are fixed reference values of $n_L$ loads and $n_R$ renewable sources; $\eta^{i}$ and $\nu^{j}$ are a random variables that follow a required distribution. Further, we will consider log-normal distribution. As in~\cite{donnot2019deep}, we separate random variables $\eta^{i}$ and $\nu^{j}$ into two components:
\begin{subequations}\label{eq:power_coef}
\begin{align}
 \eta^i = \eta_{corr} \eta^i_{uncorr}, ~\forall 1\leq i \leq n_L \\
 \nu^j = \nu_{corr} \nu^j_{uncorr}, ~\forall 1\leq j \leq n_R
\end{align}
\end{subequations}
where $\eta_{corr}$ and $\nu_{corr}$ denotes spatio-temporal correlations, while $\eta^j_{uncorr}$ and $\nu^j_{uncorr}$ are local variations. We propose that both variables have \textit{log-normal} distribution. We consider correlated variables $\eta_{corr}$ at all loads and $\nu_{corr}$ at all RES to take the same values with mean and standard deviation, ($\mu(\eta_{corr}) = -1$, $\mu(\nu_{corr}) = 0.2$) and ($\sigma(\eta_{corr}) =0.1$, $\sigma(\nu_{corr}) =0.4$). In contrast, uncorrelated variables are different for each load and renewable source, distributed with means ($\mu(\eta_{corr}) = 1$, $\mu(\nu_{corr}) = 1$) and standard deviations ($\sigma(\eta_{corr}) =0.05$, $\sigma(\nu_{corr}) =0.3$). 
 
\subsubsection{Sampling reactive power loads and renewable sources}
Reactive powers are sampled as a fixed power ratio from the active injections as follows:
\begin{subequations}
\begin{align}
 q_l^i = \gamma^i p_{l}^i, ~\forall 1\leq i \leq n_L,~~ 
 q_{rs}^j = \gamma^j p_{rs}^j, ~\forall 1\leq j \leq n_R \nonumber
\end{align}
\end{subequations}
The power ratio is derived form the reference values $\gamma^i = \frac{q_{l_{ref}}^i}{p_{l_{ref}}^i}$,~ $\gamma^j = \frac{q_{rs_{ref}}^j}{q_{rs_{ref}}^j}$.
\subsubsection{Sampling active power generation}
Since the voltages of the generation at PV nodes are fixed to their nominal values, we only considered the sampling of the active power generation $p_g$. To sample generation correctly, we have to ensure that the quasi-stationary power system equation has power balance:
\begin{equation}\label{eq:los}
 \sum_{k=1}^{n_u} p^{k}_g = \sum_{i=1}^{n_L} p^{i}_l - \sum_{j=1}^{n_R} p^{j}_{rs} + losses\\ 
\end{equation}
Consequently, the total production is approximately some percentage $\varrho$ higher than the total demand, covering losses in the power system. In our case, this percentage $\varrho$ is calculated as the ratio of the referent values between the total production and the total load for the various systems\footnote{For tested IEEE 9 bus system $\varrho$=1.0139 that means the total losses are 1.39\% of the total output. Similarly, for IEEE 39 bus system $\varrho$=1.0086}. Following~\cite{donnot2019deep}, we first formulate the part in which each production sample has to target the total demand increased by losses percentage as:
\begin{gather*}
 \Breve{p}_g^{k} = \psi^k \cdot \varrho \cdot \frac{\sum_{i=1}^{n_L} p^{i}_l}{\sum_{k=1}^{n_u} p^{k}_{g_{ref}}} \cdot p^{k}_{g_{ref}} 
\end{gather*}
where $\psi^k \sim \mathcal{U}[0.8,1.2]$ is the uniform distribution to ensure no deterministic behavior for the productions.

We applied the second step to ensure that the relation $\sum_{k=1}^{n_u} p^{k}_g + \sum_{j=1}^{n_R} p^{j}_{rs} = \varrho \cdot \sum_{i=1}^{n_L} p^{i}_l$ holds perfectly, as:
\begin{gather*}
 p_g^{k} = \Breve{p}_g^{k} \cdot \varrho \cdot \frac{\sum_{i=1}^{n_L} p^{i}_l}{\sum_{k=1}^{n_u} \Breve{p}^{k}_g} 
\end{gather*}

\section{Empirical Evaluation}\label{sec:30-Case}
The results presented in this paper are obtained on an Intel Core i7-5500U CPU @ 2.40GHz and 8GB of RAM. To simulate the experiment, we use our python gp-ccopf\footnote{\url{https://github.com/mile888/gp_cc-opf}} framework developed for this article and pandapower~\cite{thurner2018pandapower} package to validate results.

\subsection{Case Study}
We use the 9 and 39-bus IEEE Test Systems to evaluate the performance and scalability of the proposed GP CC-OPF method. Both systems are provided with pandapower~\cite{thurner2018pandapower}.

\textit{IEEE 9-bus Test Systems} consists of 3 generators and 3 loads on the high-voltage level of $V_{n} = 345kV$. We introduce two renewable source generators at buses 3 and 5 with the total forecast power output of 80 MW, which is approximately 25.4\% of the total active power demand. We assume that the power ratio is 0.3. All loads and renewable sources are uncertain.

\textit{IEEE 39-bus Test Systems} consists of 10 generators and 21 loads with $V_{n} = 345kV$. Renewable source generators are placed at buses 1, 11, 14, 21, 23, and 28 with the total forecast power output of 1260 MW, which is approximately 20.2\% of the total active power demand. The assumed power ratio is 0.3. Also, we consider that all loads and renewable sources are uncertain.

The forecast errors are modeled as zero mean, multivariate Gaussian random variables with a standard deviation corresponding to 15\% of the forecasted loads ($\sigma_l = 0.15\, p_l$) and 30\% of the forecasted renewable sources ($\sigma_r = 0.3\, p_r$). The acceptable violation probabilities $\varepsilon$ are set to $\varepsilon_u=0.1\%$ and $\varepsilon_y = 2.5\%$. Since generation constraints are physically impossible to violate, we assume a very small percentage of violations ($\varepsilon_u$), as proposed in Swissgrid (the swiss system operator)~\cite{abbaspourtorbati2015swiss} for their reserve procurement process. In contrast, output constraints are soft constraints where some violations can be tolerated ($\varepsilon_y$) if the magnitude and duration are not too large or removed through additional control actions such as generation re-dispatch~\cite{roald2017chance}.

\subsection{Evaluation Procedure and Metrics}
To evaluate the prediction quality of the \emph{GP regression}, optimized using training data, we use the root mean squared error (RMSE) metric for each output variable $a$ as:
\begin{gather*}
 RMSE_a = \sqrt{\frac{1}{N_*}\sum_{i=1}^{N_*} ([y]_{i,a} - [\hat{y}]_{i,a})^2}
\end{gather*} 
where $y$ and $\hat{y}$ are the actual (AC-PF) and predicted (GP) mean values for testing data, and $N_*$ is a number of test samples. The average evaluation score of all output variables is calculated as follows:
\begin{equation}\label{eq:rmse_1}
 RMSE = \frac{\sum_{a=1}^{n_y} RMSE_a}{n_y}
\end{equation}

Solving the data-driven CC-OPF with GP-based constraints for different test data, we compute the mean of output variables $\hat{y}$ (Eq.~\eqref{eq:balance_mean}) and report the RMSE against the variables computed by AC-PF. To analyze chance constraints feasibility, we compare the distribution of output variables, computed analytically in GP CC-OPF, with the true empirical spread computed via AC-PF with Monte-Carlo samples of uncertainty $\omega$ around the mean inputs (loads, RES, and optimized generator set-points). 

To measure the performance of the solution, we also compute the computation time, generation cost, Eq.~\eqref{eq:Gp-cost}), and the empirical spread of output variables for (a) model-based scenario CC-OPF \cite{vrakopoulou2013probabilistic,mezghani2020stochastic}, (b) AC-OPF on individual test samples (i.e., full recourse), and (c) AC-OPF for base case (mean loads and RES). Note that AC-OPF on the base case doesn't optimize for input uncertainty and presents the worst-case violations. On the other hand, the full-recourse solves AC-OPF for each sample separately and is the benchmark for feasibility as it is not restricted to linear generator feedback. 

\subsection{Models Performance}
Both GP models for IEEE 9 and 39-bus systems are trained on randomly sampled data, generated as discussed in Section \ref{sec:25-Simulation}. The GP model of the IEEE 9-bus system consists of 8 inputs features and 15 outputs, while the IEEE 39 bus has 37 inputs, 74 outputs. In the case of the IEEE 9 bus case, we applied 75 random samples for training and 25 for validation. For the IEEE 39 bus system, training is conducted on 200 samples, and 65 samples are used for validation. The average RMSE metric in Eq. \eqref{eq:rmse_1} for IEEE 9 bus and IEEE 39 bus are RMSE = $7.72e^{-5}~p.u.$ and RMSE = $8.91e^{-4}~p.u.$, respectively. The small errors suggest that the models for both systems have good generalization for deterministic inputs and that the hyperparameters are well optimized.

Next, we use the GP CC-OPF method with three approximation methods: first-order Taylor approximation (TA1), second-order Taylor approximation (TA2), and Exact moment matching (EM). The iterative solution interior-point algorithm is applied to find a feasible solution in $k$ iterations. In our simulations, the algorithm converged within a relatively small number of iterations for the convergence tolerance of $\epsilon_{tol} = 10^{-5}$, as shown in Table \ref{table:table2}. Table \ref{table:table2} also presents each approximation method's RMSE for GP-output against AC-PF based outputs, for same inputs. The RMSE results are the same for TA1 and TA2 algorithms since both algorithms have the same mean function \eqref{eq:mean_TA1}. In contrast, the EM mean function depends on input covariance; in our case, the RMSE is better. However, the CPU time of optimization is greater for EM, with a sharper increase for the larger system.

\begin{table}[ht]
\caption{GP CC-OPF method results for IEEE 9 and IEEE 39 bus systems} 
\centering 
\begin{tabular}{c | c | c c c } 
\hline\hline 

System & parameters & TA1 & TA2 & EM\\ [0.5ex] 
\hline 
\multirow{2}{*}{IEEE 9} & RMSE [p.u.] & $6.35 \cdot 10^{-3}$ & $6.35 \cdot 10^{-3}$ & $5.95\cdot 10^{-3}$\\\cline{2-5}
 & No. iteration & 16 & 14 & 16\\\cline{2-5}
 & CPU time [s] & 0.35 & 4.97 & 6.60\\\hline 
\multirow{3}{*}{IEEE 39} & RMSE [p.u.] & $1.37 \cdot 10^{-2}$ & $1.37 \cdot 10^{-2}$ & $1.35 \cdot 10^{-2}$\\\cline{2-5}
 & No. iteration & 18 & 18 & 20 \\\cline{2-5}
 & CPU time [s]& 19.38 & 717.6 & 1221.2 \\\cline{2-5}
 
\hline\hline 
\end{tabular}
\label{table:table2} 
\end{table}

\subsection{Chance-constraint feasibility and uncertainty margins}
\begin{figure}[t]
 \begin{center}
\includegraphics[width=\linewidth]{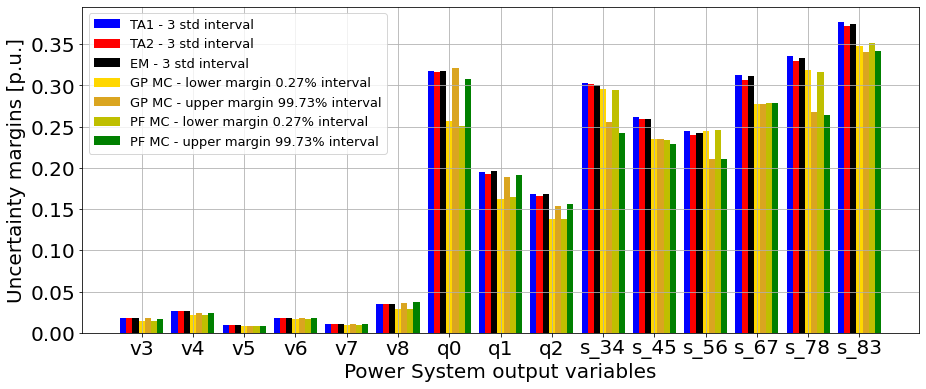}
 \caption{Output uncertainty spread of IEEE 9.}\label{fig:waveforms}
 \end{center}
\end{figure}

We plot the spread of centered CC-OPF output variables (voltages at PQ buses and reactive powers at controllable generators) for the IEEE 9-bus system in Fig.~\ref{fig:waveforms}. The three standard deviations (3 STD) intervals for uncertainty margins are analytically derived using the estimated GP-variance,  see Eq.~\eqref{eq:balance_var}. The different approximations considered for uncertainty propagation are marked as  blue (TA1), red (TA2), and black (EM). The empirical spread for the same output variables is derived using both GP-function and AC-PF equation with 5000 Monte Carlo (MC) samples of the input uncertainty $\omega$. Brown and yellow bars represent the upper and lower margins of the MC-based GP model, while dark and light green bars are for upper and lower margins of the MC-based AC-PF equation. Due to the asymmetrical MC-based distributions owing to nonlinear AC-PF, the upper and lower uncertainty margins, computed as 99.73\% and 0.27\% probability, respectively (3 STD), are depicted separately. 

\begin{figure}[!t]
 \centering
 \subfloat[\centering Voltage output distributions]{{\includegraphics[width=.95\linewidth]{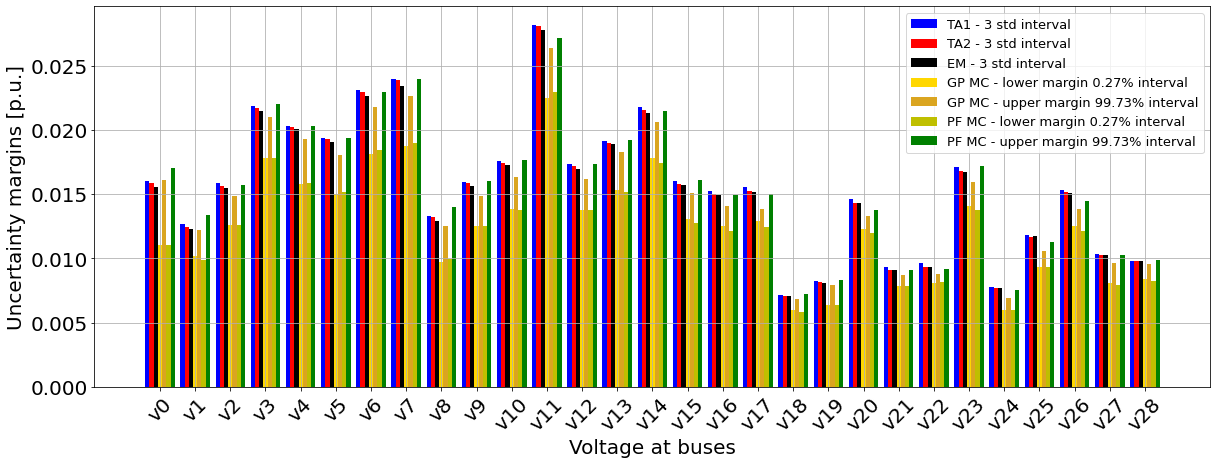} }}%
 \qquad
 \subfloat[\centering Reactive power output distributions]{{\includegraphics[width=.95\linewidth]{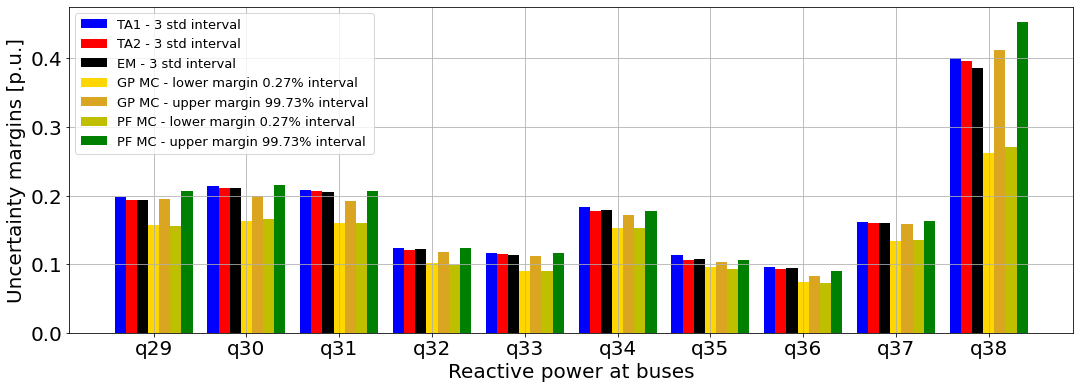} }}%
 \qquad
 \subfloat[\centering Apparent power output distributions]{{\includegraphics[width=.95\linewidth]{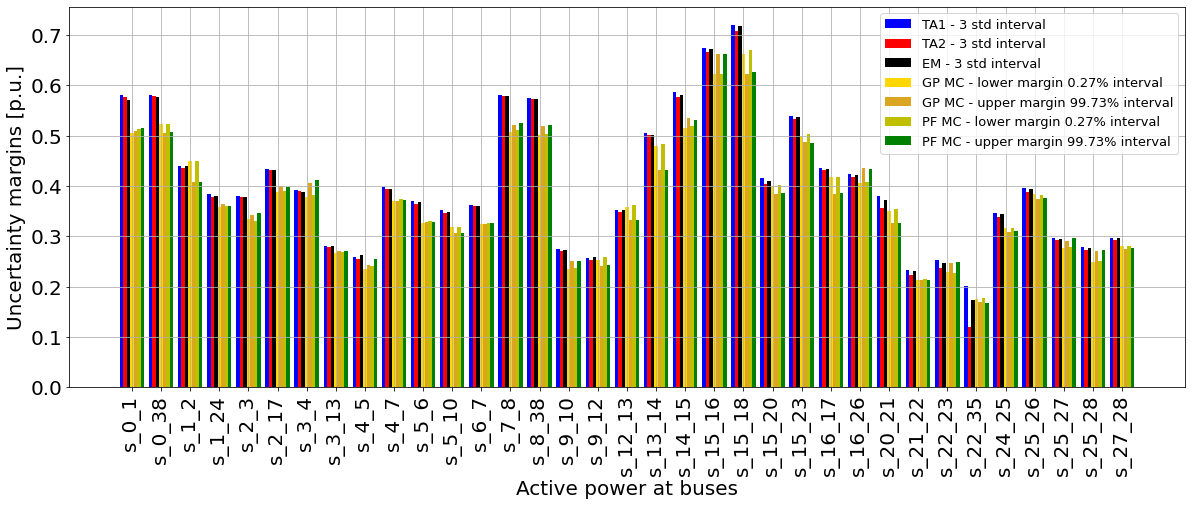} }}%
 \caption{Output uncertainty spread of IEEE 39.}%
 \label{fig:example_39}%
\end{figure}

Similar plots for nodal voltage, reactive injections, and apparent powers on lines in the 39 bus systems are presented in Fig.~\ref{fig:example_39}. Comparing the analytical with the Monte-Carlo based uncertainty margins, it is clear that they are close for most outputs, with a few exceptions where the analytical spread \emph{overestimate} the MC ones. 

Note that as analytical margins satisfy the chance constraints, see Eq.~\eqref{eq:const_var1} and Eq.~\eqref{eq:const_var2}, the smaller MC empirical violations will also be within the chance-constrained limits, and hence feasible. Moreover, from Fig.~\ref{fig:waveforms} and Fig.~\ref{fig:example_39}, MC margins for GP model and AC-PF equation are comparable, which implies that the GP model can learn the nonlinear AC-PF accurately. For the IEEE 39 bus system, we do observe small deviations between upper and/or lower margins for some variables, but this can be overcome by increasing the number of training data samples.

\subsection{Comparison with Model-based Scenario CC-OPF}
Here, we compare cost, computational time, and CC-feasibility of GP CC-OPF method (with TA1 approximation), with model-based scenario CC-OPF \cite{vrakopoulou2013probabilistic,mezghani2020stochastic}, with a varying number of scenarios. The empirical constraint violation for determined set-points is computed using $1000$ test MC samples. To benchmark, we also compare against two model-based setups:
\begin{itemize}
    \item[$(A)$] AC-OPF will full recourse, i.e., separate AC-OPF for each MC sample, 
    \item[$(B)$] and AC-OPF for the base case, i.e., AC-OPF for the mean input.
\end{itemize}

Note that though unrealistic, (A) considers generalized nonlinear feedback, while (B) doesn't optimize the participation factor $\alpha$ and uses equal generator participation for test MC samples. Hence they represent, respectively, model-based solutions with the best and worst feasible solutions. 

In Table \ref{table:table4}, we can observe that GP CC-OPF solution produces the desired reliability level of $\varepsilon_y = 2.5\%$, with a cost between that of (A) and (B) for both test cases. Note that (A) solves separate AC-OPF for each MC sample and hence has no constraint violation but higher cost. On the other hand, (B) 's set point is optimized by OPF for the mean loads and RES. Thus, it has a lower cost but large violations for test uncertainty. Their computational time is that of a single AC-OPF. For scenario CC-OPF, increasing the number of scenarios increases cost and reduces constraint violations for either test case. Although the number of scenarios necessary to get desired CC-feasibility is greater than $100$ for IEEE 39, and greater than which leads to a significantly higher computational time for scenario CC-OPF over GP CC-OPF where CC are implemented analytically. 

\begin{figure}[ht] 
 \centering
 \subfloat[Spread of voltages for GP CC-OPF and (B) AC-OPF for base case.]{
\includegraphics[width=.48\linewidth]{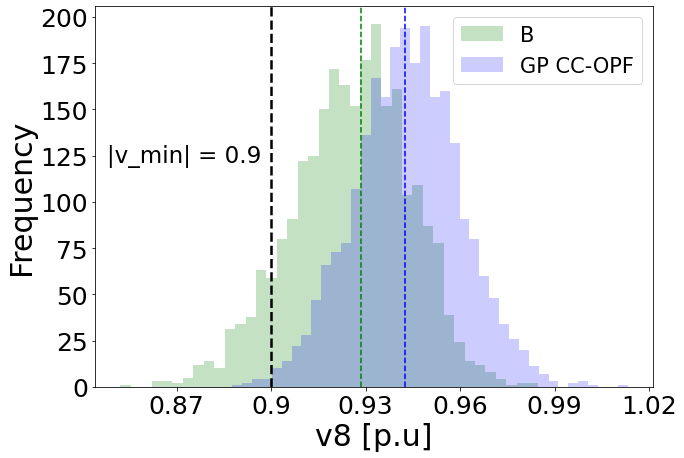}%
  \label{fig:histcase9a}%
  }%
 \hfill%
 \subfloat[Spread of voltages for scenario CC-OPF and (A) AC-OPF with full recourse.]{ \includegraphics[width=0.48\linewidth]{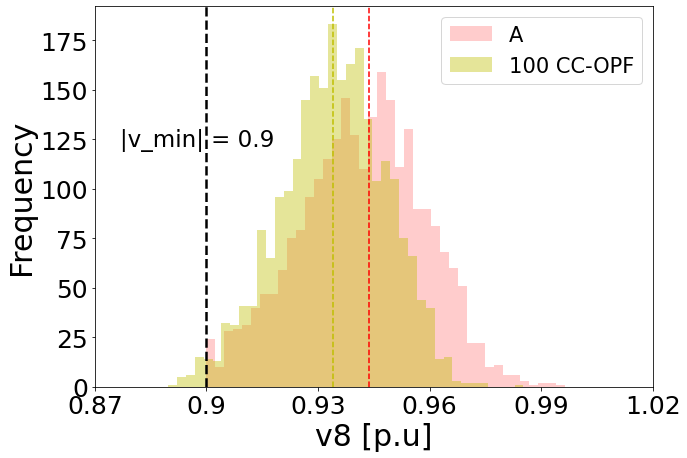}
  \label{fig:histcase9b}%
  }%
 \caption{Case IEEE-9. Empirical Spread of voltages at Bus 8 under uncertainty.}
 \label{fig:histcase9}%
\end{figure}

\begin{figure}[ht] 
 \centering
 \subfloat[Spread of voltages for GP CC-OPF and (B) AC-OPF for the base case.]{%
\includegraphics[width=0.48\linewidth]{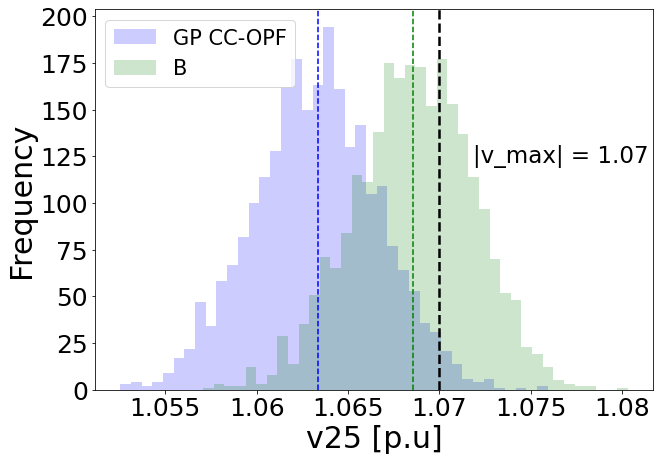}%
  \label{fig:histcase39a}%
  }%
 \hfill%
 \subfloat[The spread of voltages for scenario CC-OPF and (A) AC-OPF with full recourse.]{%
\includegraphics[width=0.48\linewidth]{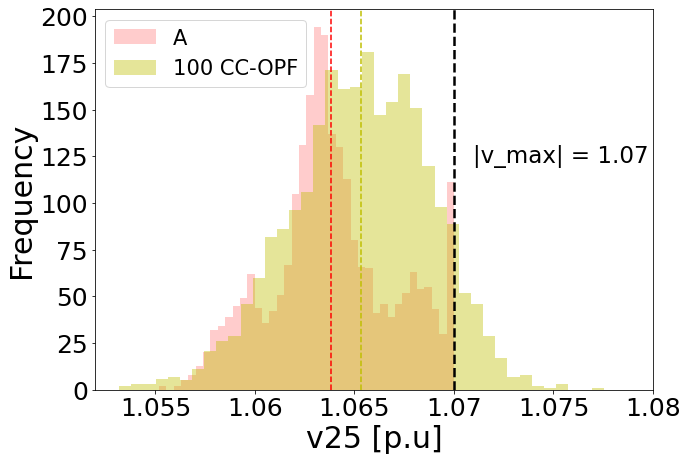}
  \label{fig:histcase39b}%
  }%
 \caption{Case IEEE-39. Empirical Spread of voltages at Bus 25 under uncertainty.}
 \label{fig:histcase39}%
\end{figure}

To study the empirical spread of output variables, we illustrate the voltage under uncertainty at bus 8 of IEEE9 in Fig.~\ref{fig:histcase9}, and voltage at bus 25 of IEEE39 in Fig.~\ref{fig:histcase39}. As one can observe, voltage values for GP CC-OPF for the IEEE9 (Fig.~\ref{fig:histcase9a}) are centered further from the minimal voltage than the corresponding voltage values for 100 scenario CC-OPF (Fig.~\ref{fig:histcase9b}). The latter leads to reduced violation of GP CC-OPF compared to scenario CC-OPF for the important lower voltage limit. Moreover, the latter's spread is more skewed to the constraint border. Note that (A) with full recourse has no violation while (B) for base-case has a higher violation. Similarly, for voltage at bus 25 of IEEE39, the GP CC-OPF voltage spread (Fig.~\ref{fig:histcase39a}) is further from the important upper limit compared to the solution of scenario CC-OPF (Fig.~\ref{fig:histcase39b}).
\begin{table}[ht]
\centering
 \begin{tabular}{l|l|l|l}
 \hline\hline
 System &
 \multicolumn{3}{c}{IEEE 39}\\\cline{2-4}
 \hline
 Parameters & Cost [\$] & Infeas. Prob. [\%] & Time (s)\\
 \hline
 A (full recourse) & $7.869\cdot 10^6$ & - & 0.93\\
 \hline
 B (base case)& $7.533 \cdot 10^6$ & 35.36 & 0.93 \\
 \hline
 GP CC-OPF & $7.752\cdot 10^6$ & 2.36 & 19.38 \\
 \hline
 50 CC-OPF & $7.642\cdot 10^6$ & 14.20 & 96.4 \\
 \hline
 100 CC-OPF & $7.696\cdot 10^6$ & 8.24 & 184.7 \\
 \hline
 200 CC-OPF& $7.813\cdot 10^6$ & 0.16 & 505.0 \\
 \hline\hline
 \end{tabular}
 \begin{tabular}{l|l|l|l}
 \hline\hline
 System &
 \multicolumn{3}{c}{IEEE 9} \\\cline{2-4}
 \hline
 Parameters & Cost [\$] & Prob. [\%] & Time (s)\\
 \hline
 A (full recourse)& $4.056\cdot 10^3$ & - & 0.86\\
 \hline
 B (base-case)& $3.467 \cdot 10^3$ & 9.76 & 0.86 \\
 \hline
 GP CC-OPF & $4.039 \cdot 10^3$ & 0.44 & 0.35\\
 \hline
 20 CC-OPF & $3.481 \cdot 10^3$ & 7.52 & 4.4\\
 \hline
 50 CC-OPF & $3.836 \cdot 10^3$ & 4.92 & 12.1\\
 \hline
 100 CC-OPF & $3.985 \cdot 10^3$ & 1.24 & 22.8\\
 \hline\hline
 \end{tabular}
 \caption{Cost function values and probability of violation of a constraint at a solution} 
 \label{table:table4}
\end{table}

\subsection{Discussion}\label{sec:40-Discussion}
Comparing the results of different approximation methods within GP CC-OPF, we can conclude that all three methods (TA1, TA2, and EM) approximate uncertainty margins very well with small probability violations for MC test samples for both systems (Figs.~\ref{fig:waveforms},~\ref{fig:example_39}). Table \ref{table:table2} shows that with the system scaling, the accuracy (RMSE) of the mean performance decreases. It can be explained by the fact that for the IEEE 39 bus system, we considered 200 training data samples randomly sampled in the space. More samples and active learning techniques can help improve the accuracy. In general, the problem of GP is scalability, where the computational complexity of training and prediction increases with the number of samples and input/output dimensions. Due to the inclusion of the second derivative term, TA2 and EM are more complicated than TA1 as confirmed by the CPU time results in Table \ref{table:table2}. Computational complexity problems can be overcome by using sparse GP and dimensionality reduction techniques \cite{hewing2019cautious}, which we plan to explore in the future. Generally, from the obtained results that the TA1 method gives similar performance accuracy as TA2 and EM, but with less complexity. 

\section{Conclusion}\label{sec:50-Conclusion}
In this research, we have proposed a data-driven and model-free stochastic OPF method by replacing the standard AC power flow equation and security constraints with the Gaussian process regression model. We show that a relatively small number of samples spread correctly in the space can generalize the GP function to model the AC-PF equations and furthermore enables accurate propagation of uncertainty from input to output. 

We show that applying the first Taylor Approximation method in GP CC-OPF outperforms other proposed approximation methods and two widely used sample-based approaches in our experiments. Although the accuracy of the data-driven CC-OPF is notable, it depends on the propagation space of training data. In future work, we will attempt to overcome this issue by proposing a new hybrid GP CC-OPF method that combines linear and non-linear approximation methods. Also, our further research will be aimed at improving the scalability of the proposed approach using sparse GP formulations. 
 \bibliographystyle{elsarticle-num} 
 \bibliography{cas-refs}

\begin{thebibliography}{10}
\expandafter\ifx\csname url\endcsname\relax
  \def\url#1{\texttt{#1}}\fi
\expandafter\ifx\csname urlprefix\endcsname\relax\def\urlprefix{URL }\fi
\expandafter\ifx\csname href\endcsname\relax
  \def\href#1#2{#2} \def\path#1{#1}\fi

\bibitem{liu2012challenges}
Z.~Liu, L.~v.~d. Sluis, W.~Winter, Challenges, experiences and possible
  solutions in transmission system operation with large wind integration,
  {OSTI} (2012).

\bibitem{stott2012optimal}
B.~Stott, O.~Alsa{\c{c}}, Optimal power flow: Basic requirements for real-life
  problems and their solutions, in: SEPOPE XII Symposium, Rio de Janeiro,
  Brazil, Vol.~11, 2012, pp. 1--10.

\bibitem{venzke2020chance}
A.~Venzke, L.~Halilba{\v{s}}i{\'c}, A.~Barr{\'e}, L.~Roald,
  S.~Chatzivasileiadis, Chance-constrained ac optimal power flow integrating
  hvdc lines and controllability, International Journal of Electrical Power \&
  Energy Systems 116 (2020) 105522.

\bibitem{roald2013analytical}
L.~Roald, F.~Oldewurtel, T.~Krause, G.~Andersson, Analytical reformulation of
  security constrained optimal power flow with probabilistic constraints, in:
  2013 IEEE Grenoble Conference, IEEE, 2013, pp. 1--6.

\bibitem{roald2017chance}
L.~Roald, G.~Andersson, Chance-constrained {AC} optimal power flow:
  Reformulations and efficient algorithms, IEEE Transactions on Power Systems
  33~(3) (2017) 2906--2918.

\bibitem{viafora2020chance}
N.~Viafora, S.~Delikaraoglou, P.~Pinson, J.~Holb{\o}ll, Chance-constrained
  optimal power flow with non-parametric probability distributions of dynamic
  line ratings, International Journal of Electrical Power \& Energy Systems 114
  (2020) 105389.

\bibitem{calafiore2006scenario}
G.~C. Calafiore, M.~C. Campi, The scenario approach to robust control design,
  IEEE Transactions on Automatic Control 51~(5) (2006) 742--753.

\bibitem{vrakopoulou2013probabilistic}
M.~Vrakopoulou, K.~Margellos, J.~Lygeros, G.~Andersson, A probabilistic
  framework for reserve scheduling and security assessment of systems with high
  wind power penetration, IEEE Transactions on Power Systems 28~(4) (2013)
  3885--3896.

\bibitem{mezghani2020stochastic}
I.~Mezghani, S.~Misra, D.~Deka, Stochastic ac optimal power flow: A data-driven
  approach, Electric Power Systems Research 189 (2020) 106567.

\bibitem{capitanescu2012cautious}
F.~Capitanescu, S.~Fliscounakis, P.~Panciatici, L.~Wehenkel, Cautious operation
  planning under uncertainties, IEEE Transactions on Power Systems 27~(4)
  (2012) 1859--1869.

\bibitem{owen2019importance}
A.~B. Owen, Y.~Maximov, M.~Chertkov, Importance sampling the union of rare
  events with an application to power systems analysis, Electronic Journal of
  Statistics 13~(1) (2019) 231--254.

\bibitem{lukashevich2021importance}
A.~Lukashevich, V.~Gorchakov, P.~Vorobev, D.~Deka, Y.~Maximov, Importance
  sampling approach to chance-constrained dc optimal power flow, arXiv preprint
  arXiv:2111.11729 (2021).

\bibitem{lukashevich2021power}
A.~Lukashevich, Y.~Maximov, Power grid reliability estimation via adaptive
  importance sampling, IEEE Control Systems Letters 6 (2021) 1010--1015.

\bibitem{bienstock2014chance}
D.~Bienstock, M.~Chertkov, S.~Harnett, Chance-constrained optimal power flow:
  Risk-aware network control under uncertainty, Siam Review 56~(3) (2014)
  461--495.

\bibitem{lubin2019chance}
M.~Lubin, Y.~Dvorkin, L.~Roald, Chance constraints for improving the security
  of ac optimal power flow, IEEE Transactions on Power Systems 34~(3) (2019)
  1908--1917.

\bibitem{lubin2015robust}
M.~Lubin, Y.~Dvorkin, S.~Backhaus, A robust approach to chance constrained
  optimal power flow with renewable generation, IEEE Transactions on Power
  Systems 31~(5) (2015) 3840--3849.

\bibitem{roald2016corrective}
L.~Roald, S.~Misra, T.~Krause, G.~Andersson, Corrective control to handle
  forecast uncertainty: A chance constrained optimal power flow, IEEE
  Transactions on Power Systems 32~(2) (2016) 1626--1637.

\bibitem{du2021chance}
X.~Du, X.~Lin, Z.~Peng, S.~Peng, J.~Tang, W.~Li, Chance-constrained optimal
  power flow based on a linearized network model, International Journal of
  Electrical Power \& Energy Systems 130 (2021) 106890.

\bibitem{muhlpfordt2019chance}
T.~M{\"u}hlpfordt, L.~Roald, V.~Hagenmeyer, T.~Faulwasser, S.~Misra,
  Chance-constrained {AC} optimal power flow: A polynomial chaos approach, IEEE
  Transactions on Power Systems 34~(6) (2019).

\bibitem{muhlpfordt2016solving}
T.~M{\"u}hlpfordt, T.~Faulwasser, V.~Hagenmeyer, Solving stochastic ac power
  flow via polynomial chaos expansion, in: 2016 IEEE Conference on Control
  Applications (CCA), IEEE, 2016, pp. 70--76.

\bibitem{dudley2010sample}
R.~M. Dudley, Sample functions of the gaussian process, in: Selected Works of
  RM Dudley, Springer, 2010, pp. 187--224.

\bibitem{schulz2018tutorial}
E.~Schulz, M.~Speekenbrink, A.~Krause, A tutorial on gaussian process
  regression: Modelling, exploring, and exploiting functions, Journal of
  Mathematical Psychology 85 (2018) 1--16.

\bibitem{liu2020gaussian}
H.~Liu, Y.-S. Ong, X.~Shen, J.~Cai, When gaussian process meets big data: A
  review of scalable gps, IEEE transactions on neural networks and learning
  systems 31~(11) (2020) 4405--4423.

\bibitem{pareek2020gaussian}
P.~Pareek, H.~D. Nguyen, Gaussian process learning-based probabilistic optimal
  power flow, IEEE Transactions on Power Systems 36~(1) (2020) 541--544.

\bibitem{zamzam2020learning}
A.~S. Zamzam, K.~Baker, Learning optimal solutions for extremely fast ac
  optimal power flow, in: 2020 IEEE International Conference on Communications,
  Control, and Computing Technologies for Smart Grids (SmartGridComm), IEEE,
  2020, pp. 1--6.

\bibitem{pan2020deepopf}
X.~Pan, T.~Zhao, M.~Chen, S.~Zhang, Deepopf: A deep neural network approach for
  security-constrained dc optimal power flow, IEEE Transactions on Power
  Systems 36~(3) (2020) 1725--1735.

\bibitem{kekatos}
M.~K. Singh, V.~Kekatos, G.~B. Giannakis, Learning to solve the {AC-OPF} using
  sensitivity-informed deep neural networks, IEEE Transactions on Power Systems
  (2021).

\bibitem{gupta2021dnn}
S.~Gupta, S.~Misra, D.~Deka, V.~Kekatos, Dnn-based policies for stochastic ac
  opf, arXiv preprint arXiv:2112.02441 (2021).

\bibitem{hewing2019cautious}
L.~Hewing, J.~Kabzan, M.~N. Zeilinger, Cautious model predictive control using
  gaussian process regression, IEEE Transactions on Control Systems Technology
  28~(6) (2019) 2736--2743.

\bibitem{liu2009theory}
B.~Liu, B.~Liu, Theory and practice of uncertain programming, Vol. 239,
  Springer, 2009.

\bibitem{apostolopoulou2014automatic}
D.~Apostolopoulou, P.~W. Sauer, A.~D. Dom{\'\i}nguez-Garc{\'\i}a, Automatic
  generation control and its implementation in real time, in: 2014 47th Hawaii
  International Conference on System Sciences, IEEE, 2014, pp. 2444--2452.

\bibitem{rasmussen2003gaussian}
C.~E. Rasmussen, Gaussian processes in machine learning, in: Summer school on
  machine learning, Springer, 2003, pp. 63--71.

\bibitem{girard2003gaussian}
A.~Girard, C.~E. Rasmussen, J.~Quinonero-Candela, R.~Murray-Smith, Gaussian
  process priors with uncertain inputs application to multiple-step ahead time
  series forecasting, in: Advances in neural information processing systems,
  Vol.~15, 2003, pp. 1--10.

\bibitem{deisenroth2010efficient}
M.~P. Deisenroth, Efficient reinforcement learning using Gaussian processes,
  Vol.~9, KIT Scientific Publishing, 2010.

\bibitem{wachter2006implementation}
A.~W{\"a}chter, L.~T. Biegler, On the implementation of an interior-point
  filter line-search algorithm for large-scale nonlinear programming,
  Mathematical programming 106~(1) (2006) 25--57.

\bibitem{andersson2019casadi}
J.~A. Andersson, J.~Gillis, G.~Horn, J.~B. Rawlings, M.~Diehl, Casadi: a
  software framework for nonlinear optimization and optimal control,
  Mathematical Programming Computation 11~(1) (2019) 1--36.

\bibitem{donnot2019deep}
B.~Donnot, Deep learning methods for predicting flows in power grids: novel
  architectures and algorithms, Ph.D. thesis, Universit{\'e} Paris Saclay
  (COmUE) (2019).

\bibitem{thurner2018pandapower}
L.~Thurner, A.~Scheidler, F.~Sch{\"a}fer, J.-H. Menke, J.~Dollichon, F.~Meier,
  S.~Meinecke, M.~Braun, pandapower—an open-source python tool for convenient
  modeling, analysis, and optimization of electric power systems, IEEE
  Transactions on Power Systems 33~(6) (2018) 6510--6521.

\bibitem{abbaspourtorbati2015swiss}
F.~Abbaspourtorbati, M.~Zima, The swiss reserve market: Stochastic programming
  in practice, IEEE Transactions on Power Systems 31~(2) (2015) 1188--1194.

\end{thebibliography}





\end{document}